\documentclass[10pt,conference]{IEEEtran}

\IEEEoverridecommandlockouts
\usepackage[utf8]{inputenc}
\usepackage{graphicx}
\usepackage{url}
\usepackage{color}
\usepackage[colorlinks,allcolors=magenta]{hyperref}
\usepackage{acro}

\usepackage{amsmath}
\usepackage{amsfonts}

\DeclareAcronym{ADA}{short=ADA,long=advanced driver assistance}
\DeclareAcronym{ADS}{short=ADS,long=automated driving system}
\DeclareAcronym{HMI}{short=HMI,long=human-machine interaction}
\DeclareAcronym{SA}{short=SA,long=situational awareness}
\DeclareAcronym{SOTIF}{short=SOTIF,long=safety of the intended functionality}

\title{Maintaining driver attentiveness in shared-control autonomous driving\thanks{This research was funded by the Lloyds Register Foundation under the
Assuring Autonomy International Programme grant Safe-SCAD.}}
\author{\IEEEauthorblockN{Radu Calinescu, Naif Alasmari and Mario Gleirscher}
\IEEEauthorblockA{\textit{Department of Computer Science},
\textit{University of York},
York, U.K.\\
\{radu.calinescu,nnma500,mario.gleirscher\}@york.ac.uk}}

\newcommand{\acronym}{Safe-SCAD}

\usepackage{dblfloatfix}

\newcommand{\squishlist}{
   \begin{list}{$\bullet$}
    { \setlength{\itemsep}{2pt}%
    \setlength{\parsep}{0pt}
      \setlength{\topsep}{5pt}     \setlength{\partopsep}{0pt}
      \setlength{\leftmargin}{1.15em} \setlength{\labelwidth}{0.8em}
      \setlength{\labelsep}{0.5em} } }

\newcommand{\squishlisttwo}{
   \begin{list}{$\bullet$}
    { \setlength{\itemsep}{0pt}    \setlength{\parsep}{0pt}
      \setlength{\topsep}{0pt}     \setlength{\partopsep}{0pt}
      \setlength{\leftmargin}{1.35em} \setlength{\labelwidth}{1em}
      \setlength{\labelsep}{0.5em} } }

\newcommand{\squishlistBigLabel}{
   \begin{list}{$\bullet$}
    { \setlength{\itemsep}{2pt}%
    \setlength{\parsep}{0pt}
      \setlength{\topsep}{5pt}     \setlength{\partopsep}{0pt}
      \setlength{\leftmargin}{2.35em} \setlength{\labelwidth}{1.8em}
      \setlength{\labelsep}{0.5em} } }

\newcommand{\squishend}{
    \end{list}  }

\begin{document}
\maketitle 
\begin{abstract}
We present a work-in-progress approach to improving driver attentiveness in cars provided with automated driving systems. The approach is based on a control loop that \emph{monitors} the driver's biometrics (eye movement, heart rate, etc.) and the state of the car; \emph{analyses} the driver's attentiveness level  using a deep neural network; \emph{plans} driver alerts and changes in the speed of the car using a formally verified controller; and \emph{executes} this plan using actuators ranging from acoustic and visual to haptic devices. The paper presents (i)~the self-adaptive system formed by this monitor-analyse-plan-execute (MAPE) control loop, the car and the monitored driver, and (ii)~the use of probabilistic model checking to synthesise the controller for the planning step of the MAPE loop. 
\end{abstract}


\vspace*{-3mm}
\section{Introduction}

The J3016 standard~\cite{SAE-J3016-2018} classifies automated driving systems (ADSs) on a six-level scale, from no  automation at Level~0 to full automation at Level~5. Despite huge R\&D budgets and much hype over the past decade, fully autonomous (Level~5) cars are unlikely to become available to the general public any time soon. In contrast, cars providing Level 2 (i.e., partial) automation can be purchased from manufacturers including Tesla, Nissan and BMW; and  the approval of Level~3 (i.e., conditional automation) and~4 (i.e., high automation) cars is considered by regulators worldwide \cite{ALKS2020,AB2866,EU2019-2144,IMAI2019263,UNECE-2020}. 

A critical requirement for vehicles operating at autonomy Levels~2 and 3 is that a user resides in the driver's seat and is sufficiently attentive to be able to share the control of the car with the ADS. At Level~2, this human in the loop is expected to `\emph{complete the object and event detection and response subtask and} [to] \emph{supervise the driving automation system}', while at Level~3 the user is expected to be `\emph{receptive to ADS-issued requests to intervene} [\ldots] \emph{and} [to] \emph{respond appropriately}'~\cite{SAE-J3016-2018}. Although Level~4 ADSs do not rely on human support, they may still issue \emph{timely requests for human intervention} (e.g., when they approach roads or traffic situations they were not designed to handle), performing a minimum-risk manoeuvre (e.g., stopping the car safely) if their user does not respond. 

In all these scenarios, accidents with potentially fatal consequences (for Levels~2 and~3) and frequent emergency stops (for Level~4) can only be avoided if the drivers are sufficiently attentive to be able to take over the control of their vehicles~\cite{Merat2009-HowDoDrivers}. However, humans find it extremely difficult to remain attentive when overseeing the operation of automated and autonomous systems~\cite{chai2016driver,duffy2015case,matthews2017handbook}. In the automotive domain, this is amply demonstrated by accidents involving both cars with Level~2 ADS used by regular drivers~\cite{banks2018driver,TeslaAccident2018} and cars with higher autonomy levels tested by professional safety drivers~\cite{favaro2017examining}. 

Our paper proposes an approach that mitigates this problem by using a monitor-analyse-plan-execute (MAPE) control loop to improve driver attentiveness in shared-control autonomous driving. The monitoring component of this MAPE loop uses an array of sensors to collect driver biometrics and vehicle data. The analysis component uses these data and a deep neural network~\cite{CHI-2021} developed by a complementary research strand of our Safe-SCAD project\footnote{\underline{Safe}ty$\!$ of$\!$ \underline{S}hared$\!$ \underline{C}ontrol$\!$ in$\!$ \underline{A}utonomous \underline{D}riving,$\!$ \url{https://cutt.ly/Safe-SCAD}} to predict the driver response time and response quality to a potential ADS intervention request. These predictions are processed based on a simple driver attentiveness model that factors in the speed of the car, and the resulting classification of the driver as attentive, semi-attentive or inattentive guides the planning of driver alerts and car speed changes by a formally verified discrete-event controller. This controller achieves Pareto-optimal trade-offs between risk level, driver nuisance, and progress with the journey.

The first version of our \acronym\ approach is aimed at Level~3/4 ADSs, with a particular focus on the \emph{Automated Lane Keeping System} (ALKS) for which the United Nations World Forum for Harmonization of Vehicle Regulations adopted a new UN regulation~\cite{UNECE-2020} in June 2020, and that the UK Department for Transport plans to implement on UK motorways~\cite{ALKS2020}. For these advanced ADSs, the \acronym\ improvement in driver attentiveness can lead to fewer minimum-risk manoeuvres and better progress with the car journey, and with minimal risk of accidents. Future enhancements (summarised later in the paper) will mitigate additional uncertainties associated with the challenging problem tackled by our approach, extending its applicability to Level~2 ADSs, and to autonomous car testing by professional test drivers. 

The key contributions of the paper are the presentation of the driver attentiveness management problem in ALKS (as a motivating example, in Sect.~\ref{sect:problem}), the \acronym\ approach to improving driver attentiveness in shared-control autonomous driving (Sect.~\ref{sect:approach}), and the probabilistic model checking method for synthesising the \acronym\ planning component (Sect.~\ref{sect:synthesis}). The paper also discusses related work (Sect.~\ref{sec:relwork}) and summarises our plans for future work (Sect.~\ref{sect:conclusion}).

\section{Driver attentiveness management problem \label{sect:problem}}

\subsection{Background}

We consider an ADS with the characteristics stipulated in the United Nations' ALKS regulation~\cite{UNECE-2020}. The ALKS can be activated by a driver (who must be available in the driving seat, with the seatbelt fastened) when all its components are fully operational, and the vehicle is on roads and in environment (e.g., weather) conditions within its operational design domain (ODD). When activated, the system keeps the vehicle inside its lane, controlling the vehicle speed (within the range 0 to 60~km/h) to adapt to the surrounding road traffic. Additionally, the ALKS can detect the risk of collision (e.g., due to a stationary vehicle) and can stop to avoid the collision, e.g., by performing an emergency manoeuvre. 

In certain situations, all of which it must recognise, the ALKS issues a \emph{transition demand}, i.e., a request for the driver to take over the control of the vehicle. The regulation allows these situations to differ across manufacturers. However, they must include the situations in which the ALKS activation conditions are not met (e.g., the vehicle approaches a road outside its ODD), and those in which the driver is unavailable (i.e., not in the driving seat or \emph{inattentive}) and not responding to ALKS alerts aimed at restoring the driver's availability. Transition demands are issued timely, allowing (i)~an \emph{attentive driver} to resume the manual driving safely, or (ii)~the ALKS to perform a minimum-risk manoeuvre (MRM), e.g., to bring the vehicle to a standstill, if the driver is inattentive. The ALKS may reduce the vehicle speed to ensure safety, e.g., by allowing the driver additional time for the control takeover.  

It follows from the summary so far that the ALKS must be capable of assessing the driver's availability, including their position in the car (in the driving seat, wearing the seatbelt) and their attentiveness. For the latter, the  regulation proposes the use of driver biometrics such as \emph{`eye blinking, eye closure, conscious head or body movement'}~\cite{UNECE-2020}, but allows manufacturers to select their own methods for assessing driver attentiveness. Likewise, the regulation and UK's ALKS plan~\cite{ALKS2020} recommend the use of optical, acoustic and haptic warning signals (i)~to announce transition demands to the driver, and (ii)~to improve driver attentiveness, but are not prescriptive about how these alerts should be used. In the next section, we use these recommendations to define the \emph{driver attentiveness management problem} for ALKS-like ADsS.

\subsection{Problem definition \label{subsect:problem}}

Given an ALKS, we assume that its driver can have one of $n\geq 2$ attentiveness levels. The highest level (`attentive') corresponds to the situation in which the driver can respond timely to a transition demand, even at the maximum speed permitted for the vehicle. The lowest level (`inattentive') corresponds to the situation where the ALKS needs to execute an MRM unless the driver improves their level of attentiveness within a mandated time period $\tau>0$.\footnote{The UK consultation document proposes $\tau=15$s~\cite[p.\ 17]{ALKS2020}.} If present, any intermediate levels (e.g., `semi-attentive' for $n=3$) correspond to diminished driver attentiveness that does not require an MRM. However, they provide an opportunity for issuing alerts to improve the driver's attentiveness level before it drops further, and drastic action is required: MRMs  involve stopping the vehicle in a motorway lane~\cite{ALKS2020,UNECE-2020}, and should only be used as a last resort.

We assume that the ADS has two mechanisms it can use when the driver is not `attentive'. First, it can activate one or several of $m\geq 1$ \emph{alerts} (e.g., optical, acoustic and haptic) as needed to improve the driver's attentiveness. Second, it can reduce the car speed to one of $q\geq 1$ speed levels, where we allow $q=1$ for the case when this feature is not available. As such, the ALKS state at any point in time is characterised by:
\begin{enumerate}
\item the driver attentiveness level $l\in\{0,1,\ldots,n-1\}$, where $l=0$ and $l=n-1$ correspond to the driver being `attentive' and `inattentive', respectively;
\item the set of active alerts $a\in \{0,1\}^m$, where $a=(a_1,a_2,\ldots,a_m)$ indicates that the $i$-th alert is inactive when $a_i=0$, and active when $a_i=1$;
\item the vehicle speed level $v\in\{0,1,\ldots,q-1\}$.
\end{enumerate}
Using the notation $L=\{0,1,\ldots,n-1\}$, $A=\{0,1\}^m$ and $V=\{0,1,\ldots,q-1\}$ to denote the range for the three components of the ALKS state, we further assume that the following  measures are defined over the state space $L \times A\times V$:
\begin{enumerate}
\item $\mathit{nuisance} : A \rightarrow \mathbb{R}_{\geq 0}$, where $\mathit{nuisance}(a)$ represents the nuisance experienced by the driver when the alerts $a\in A$ are in use, with $\mathit{nuisance}(0,0,\ldots,0)=0$;
\item $\mathit{progress} : V\rightarrow \mathbb{R}_{\geq 0}$, where $\mathit{progress}(v)$ reflects the progress with the journey made when the vehicle travels at speed $v\in V$ (e.g., the distance travelled in one hour);
\item $\mathit{risk} : L \times V\rightarrow \mathbb{R}_{\geq 0}$, where $\mathit{risk}(l,v)$ provides a measure of the risk associated with travelling at speed $v\in V$ when the driver attentiveness level is $l\in L$,
\end{enumerate}
and that $\mathit{risk}_\mathsf{MRM}>0$ denotes the risk associated with performing an MRM. 

Finally, we assume that timing data are available about the drivers' transition between the attentiveness levels $L$, when different alert combinations are active, and at different vehicle speeds. These data may be available from studies of driver behaviour~\cite{lotz2018predicting,maia2013short}, experiments carried out by ALKS manufacturers, observations of drivers who are using the deployed ADS, or a combination thereof. Given such data, the \emph{driver attentiveness management problem} is to find a combination of alerts $a(s)\in A$ to use in each ALKS  state $s\in L \times A\times V$, such that the ALKS achieves Pareto optimality between minimising the driver nuisance, maximising the progress with the journey, and minimising the risk over a period of $T$ hours of driving.

\section{The Safe-SCAD approach \label{sect:approach}}

\begin{figure}
\centering
\includegraphics[width=0.82\hsize]{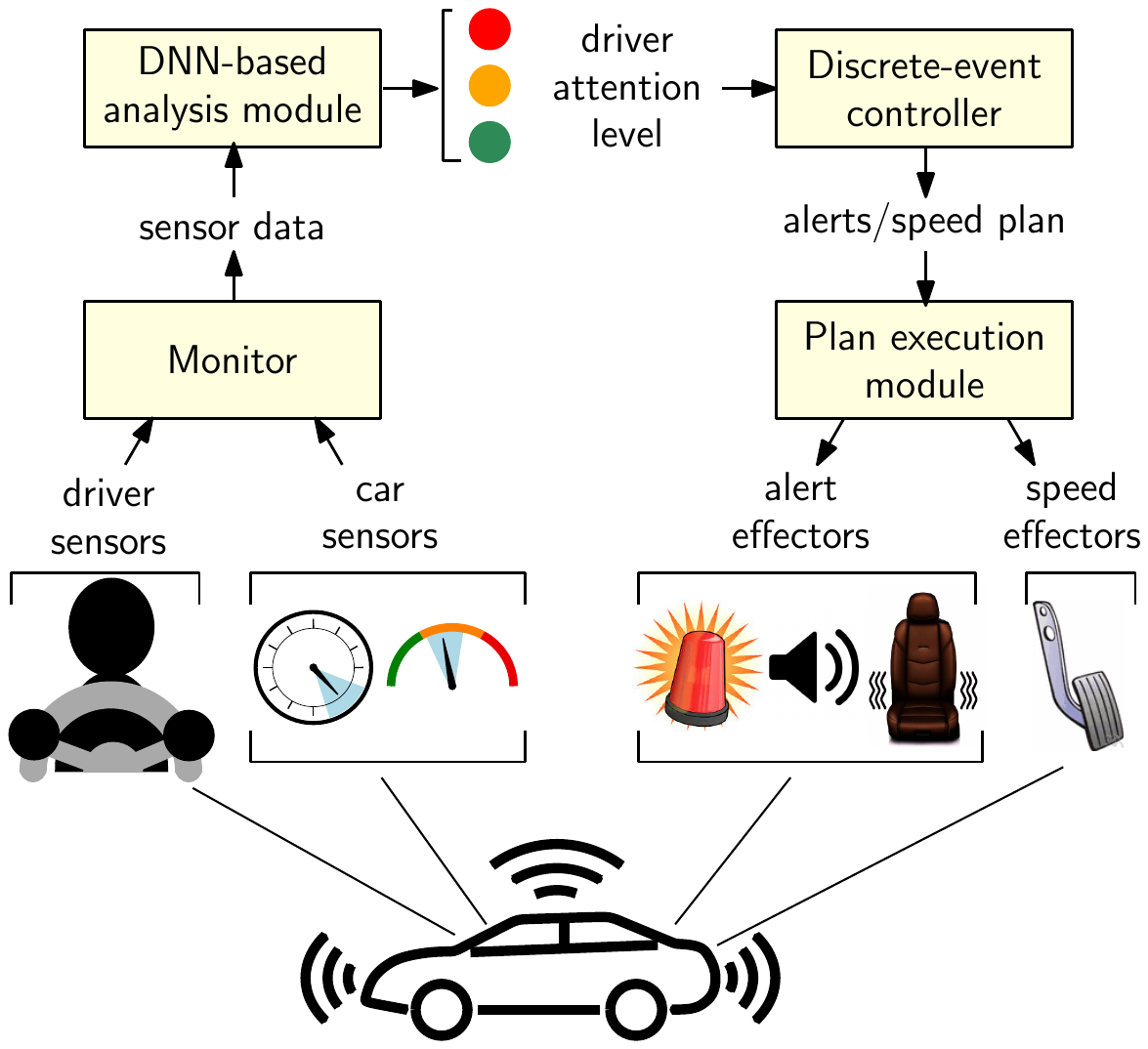}

\vspace*{-1mm}
\caption{Safe-SCAD driver attentiveness management approach for ALKS with $n=3$ attentiveness levels (attentive, semi-attentive, and inattentive) \label{fig:approach}}

\vspace*{-3mm}
\end{figure}

Our Safe-SCAD approach addresses the driver attentiveness management problem from Section~\ref{subsect:problem} by using a MAPE control loop with the components shown in Figure~\ref{fig:approach}. These components and the four stages of the MAPE loop are described in the following sections. We provide only a brief summary for the components used in the first two MAPE stages (monitoring and analysis), as their technical details are available$\!$ in$\!$ \cite{CHI-2021,Grese-et-al-2020},$\!$ and$\!$ we$\!$ focus$\!$ instead$\!$ on$\!$ their$\!$ integration into a MAPE loop and on the planning MAPE stage, which represent the two key theoretical contributions of this paper.

\subsection{Monitoring}

In this MAPE stage, data are collected from a combination of driver-biometrics sensors and vehicle sensors. In our project, driver biometrics are obtained~\cite{CHI-2021} using: (i)~eye-tracking glasses to monitor eye movement data (e.g., gaze position and fixation time); (ii)~smartwatch photoplethysmographic sensors to monitor heart rate; and (iii)~smartwatch galvanic skin response sensors to monitor hand sweating. A broad range of vehicle data streams are already collected and used by ADSs, and can easily be exploited within our MAPE loop. These range from vehicle velocity and steering wheel angle to lane position and throttle/brake pedal angles. 

\subsection{Analysis}

\begin{figure}
\centering
\includegraphics[width=\hsize]{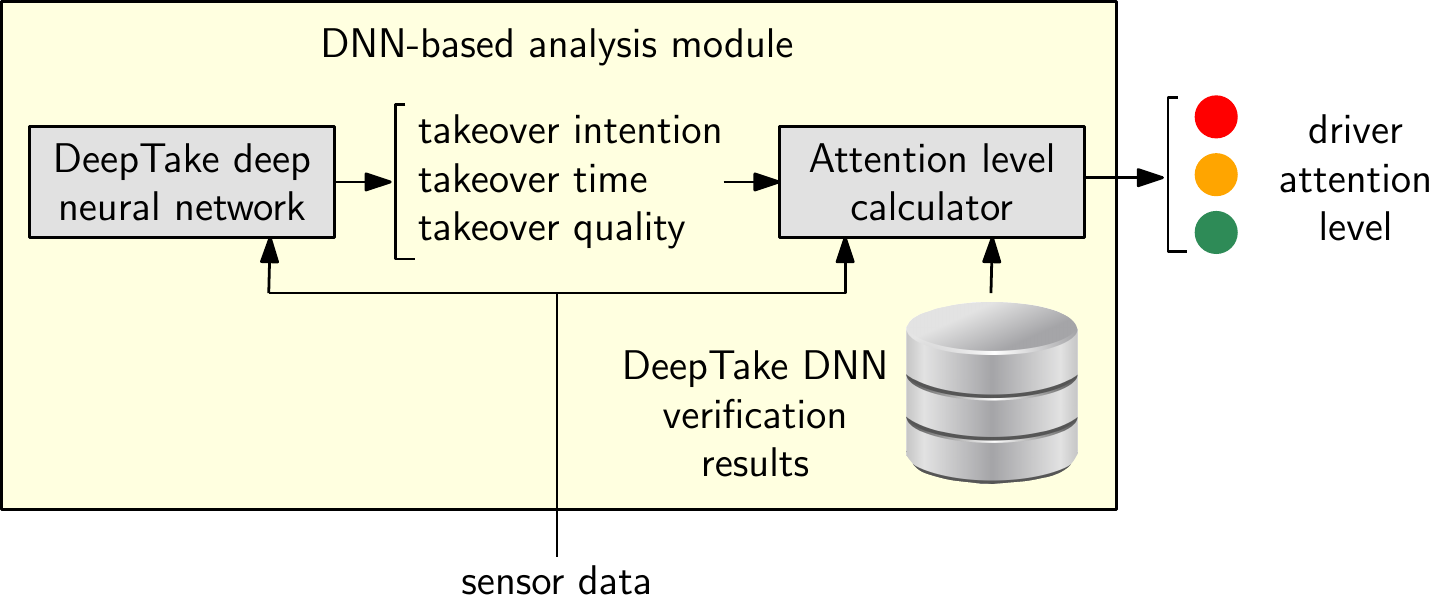}
\caption{Safe-SCAD analysis MAPE stage \label{fig:analysis}}
\end{figure}

This MAPE stage (Figure~\ref{fig:analysis}) uses the DeepTake predictor of driver takeover behaviour~\cite{CHI-2021} developed by another Safe-SCAD research strand. DeepTake is a deep neural network (DNN) that uses the driver-biometrics and vehicle data from the monitoring stage to predict the driver's control takeover:
\begin{enumerate}
    \item \emph{intention}, i.e., whether the driver would react to an ADS control-transition demand or not;
    \item \emph{time} elapsed from the transition demand until the driver assumes manual control of the vehicle, as defined by the ISO~21959 standard~\cite{ISO21959};
    \item \emph{quality} of the driver's manoeuvring of the vehicle after manual control is resumed. 
\end{enumerate} 
DeepTake was shown~\cite{CHI-2021} to predict these driver takeover metrics with an accuracy of 96\%, 93\% and 83\%, respectively. We emphasise that these accuracy levels are sufficient for the Safe-SCAD driver attentiveness management because our solution is intended for use with ALKS that can ensure safety at all times, e.g., by performing an MRM if necessary. 

\begin{figure}[b]
\centering
\includegraphics[width=0.95\hsize]{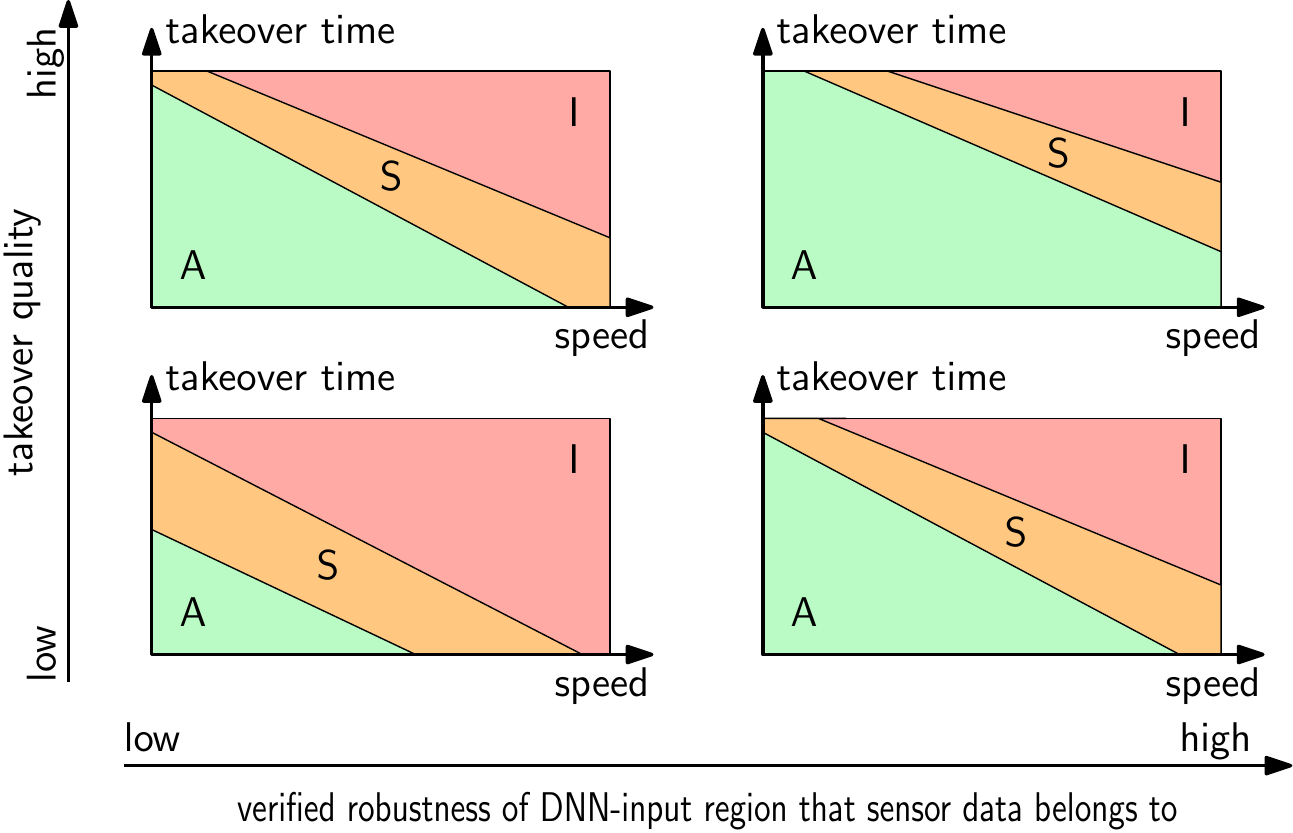}
\caption{The driver attention level (A=attentive, S=semi-attentive, I=inattentive) depends on the predicted driver takeover time and quality, on the speed of the vehicle, and on the verified robustness of the DeepTake input region that the sensor data belongs to. The diagram applies to a positive takeover intention (i.e., driver responsive to a control-transition demand); when DeepTake predicts a negative intention, the driver state is deemed inattentive. \label{fig:analysis-processing}}
\end{figure}

Additionally, our DNN-based analysis module operates conservatively by also exploiting the results from the design-time \emph{robustness verification of  DeepTake}~\cite{Grese-et-al-2020}. Figure~\ref{fig:analysis-processing} shows how we intend to use these verification results in the post-processing of the DeepTake predictions, such that the computed driver attention level is lowered when the sensor data belongs to regions of the DeepTake input space that were not identified as robust by the DNN verification. This part of our Safe-SCAD approach is under development.

\begin{figure*}[b]
\centering

\vspace*{-5mm}
\includegraphics[width=0.7\hsize]{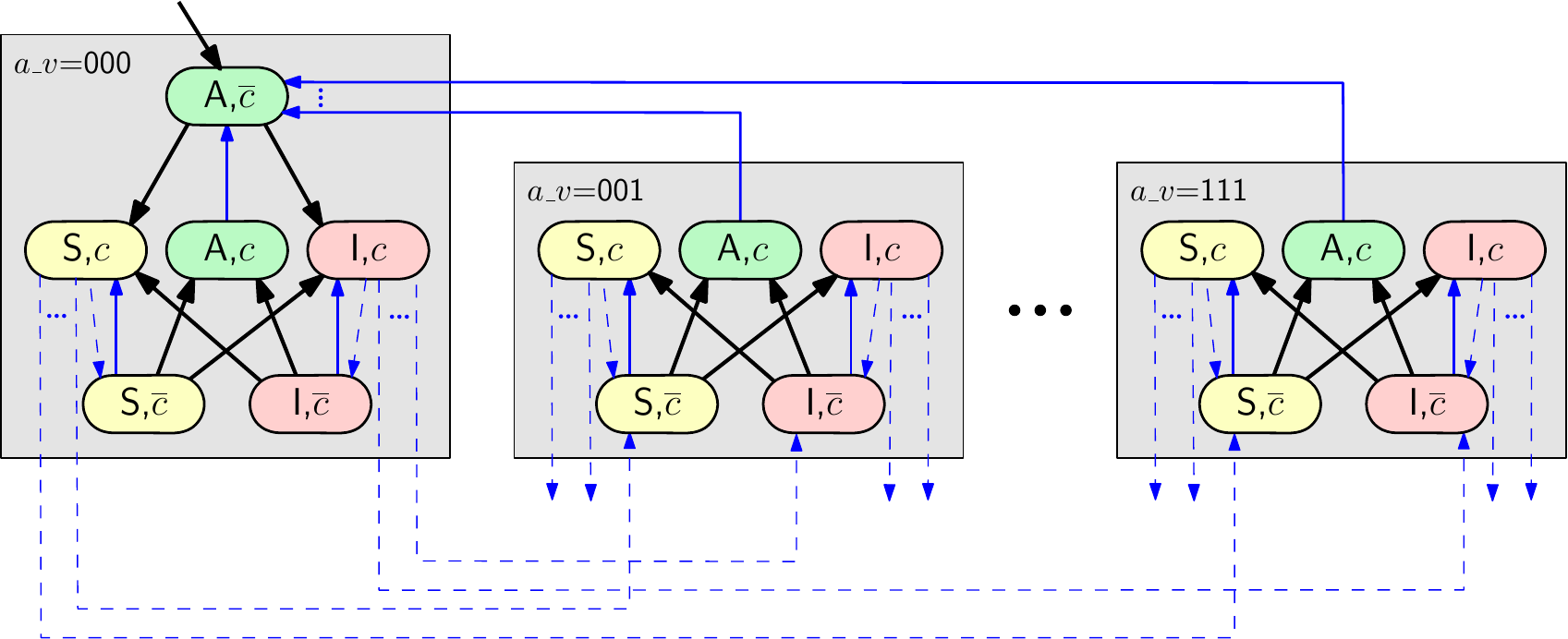}\hspace*{4mm}
\includegraphics[width=0.23\hsize]{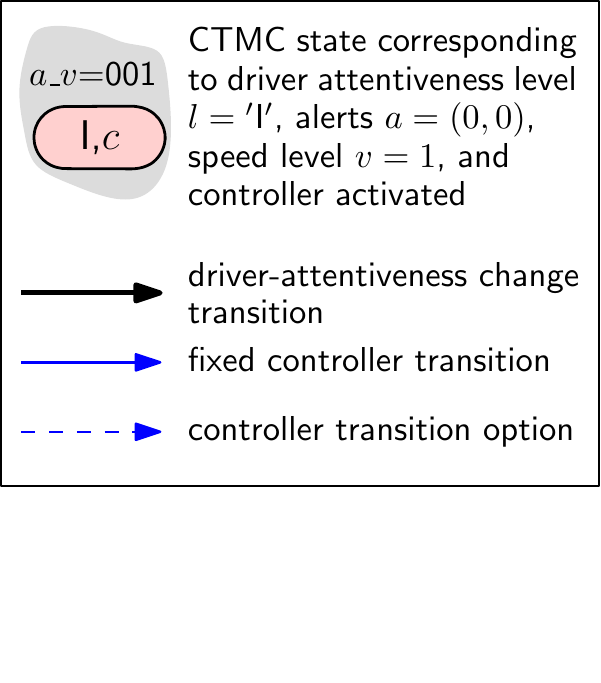}
\caption{\acronym\ controller design space for the driver attentiveness management problem with three driver attentiveness levels ($L=\{\mathsf{'A'},\mathsf{'S'},\mathsf{'I'}\}$, where $\mathsf{A}$=attentive, $\mathsf{S}$=semi-attentive and $\mathsf{I}$=inattentive), two alerts ($A=\{0,1\}^2$) and two speed levels ($V=\{0,1\}$). In the initial state (indicated by an incoming arrow) the driver is attentive, the controller is inactive, the alerts $a=(0,0)$ are inactive, and the car drives at nominal speed level $v=0$; for brevity, this combination of alert activations and speed level is denoted $a\_v=000$ in the diagram. \label{fig:CTMC}} 
\end{figure*}

\subsection{Planning}

In this MAPE stage, a discrete-event controller plans the set of active alerts $a\in A$ and the speed level $v\in V$ that the vehicle should employ. This controller is activated by the occurrence of two types of events: 
\begin{itemize}
    \item changes in the attentiveness level of the driver;
    \item the expiry of a timer that is used to activate the controller periodically at all times when the driver attentiveness level is not `attentive'.
\end{itemize}
The timer enables the controller to periodically ``try'' new or additional alerts and/or speed adjustments when the driver attentiveness level was not improved by the execution of the plan devised by the previous controller activation. The synthesis of the \acronym\ controller is detailed in Section~\ref{sect:synthesis}.

\subsection{Execution}

In this MAPE stage, the alert and speed effectors of the vehicle are used to implement the alerts and speed plan provided by the \acronym\ controller. 

\section{\acronym\ controller synthesis \label{sect:synthesis}}

To synthesise the \acronym\ controller used in the planning MAPE stage, we model the relevant behaviour of the self-adaptive system from Figure~\ref{fig:approach} as a parametric continuous-time Markov chain (CTMC). The parameters of this CTMC are chosen such that the (non-parametric) CTMC induced by each combination of parameter values corresponds to a different feasible controller, and the %
CTMCs obtained by considering all valid combinations of parameter values define the set of feasible \acronym\ controllers, i.e., the \emph{controller design space}. Given this design space, we synthesise formally verified controllers by using (i)~probabilistic model checking to determine the $\mathit{nuisance}$, $\mathit{progress}$ and $\mathit{risk}$ associated with any specific controller; and (ii)~multi-objective genetic algorithms to find controllers that achieve Pareto-optimal trade-offs between these three measures. We detail the steps of our controller synthesis process in Sections~\ref{subsect:synthesis-design-space} and \ref{subsect:synthesis-Pareto}, after a brief introduction to continuous-time Markov chains and their probabilistic model checking in Section~\ref{subsect:synthesis-CTMCs}.

\subsection{Continuous-time Markov chains \label{subsect:synthesis-CTMCs}}

A CTMC is a finite state-transition model $M=(S,s_0,\mathbf{R})$, where $S$ is a finite set of states, $s_0\in S$ is the initial state, and
$\mathbf{R}: S\times S\to [0,\infty)$ is a transition rate matrix such that for any state $s_i\in S$, the probability that the CTMC will transition from state $s_i$ to another state within $t>0$ time units is $1-e^{-t\cdot \sum_{s_k\in S} \mathbf{R}(s_i,s_k)}$, and the probability that the new state is $s_j\in S$ is given by $\mathbf{R}(s_i,s_j)/\sum_{s_k\in S} \mathbf{R}(s_i,s_k)$. 

To enable the analysis of a broader range of CTMC properties, the states and transitions of a CTMC are often  annotated with \emph{rewards}. A \emph{reward structure} over a CTMC with state set $S$ is a pair of functions $r^X=(r_1, r_2)$, where $r_1\!:\! S \!\rightarrow\! \mathbb{R}_{\geq 0}$ is a \emph{state reward function} that defines the rate $r_1(s)$ at which the reward is obtained while the Markov chain is in state $s$; and $r_2\!:\! \!S\! \times\! S \!\rightarrow\! \mathbb{R}_{\geq 0}$ is a \emph{transition reward function} that defines the reward obtained each time a transition occurs. 

Probabilistic model checkers such as PRISM~\cite{prism} and Storm~\cite{storm} use efficient symbolic model checking techniques to analyse a wide range of CTMC properties expressed in \emph{continuous stochastic logic} (CSL) augmented with rewards~\cite{KNP07a}. These properties include bounded and unbounded probabilistic reachability, and several types of reward properties. For our \acronym\ controller synthesis, we are interested in \emph{cumulative reward properties}. These properties are expressed using the CSL formula $R^X_{=?}[C^{\leq T}]$, which denotes the expected value of the reward $X$ accrued within the time interval $[0,T]$. 

\subsection{\acronym\ controller design space \label{subsect:synthesis-design-space}}

We model the driver attentiveness management problem using a family of CTMCs to define the \emph{design space} (i.e., the possible variants) of the \acronym\ controller. Each CTMC in the family has the state set $S=L\times A\times V \times \{c, \overline{c}\}$, where $L$, $A$ and $V$ are defined in Section~\ref{subsect:problem}, and $c$ is a Boolean state variable that indicates whether the discrete-event controller is active or not. As shown in Figure~\ref{fig:CTMC}, which depicts the controller design space for a specific instance of the problem, the model has three types of state transitions:

\squishlist
\item[1)] \emph{Transitions\hspace*{-0.2mm} corresponding\hspace*{-0.2mm} to\hspace*{-0.2mm} changes\hspace*{-0.2mm} in\hspace*{-0.2mm} driver\hspace*{-0.2mm} attentiveness.}
These transitions occur from states $(l,a,v,\overline{c})$, in which the controller is inactive, to states $(l',a,v,c)$ with a different driver attentiveness level (i.e., $l'\neq l$), no changes in the alerts $a$ and speed $v$, and the controller activated.

\item[2)] \emph{Transitions corresponding to fixed controller actions.} There are two classes of such actions. In the first, the CTMC transitions from states $(\mathsf{'A'},a,v,c)$, in which the driver is attentive and the controller activated, to the initial state $(\mathsf{'A'},(0,0),0,\overline{c})$;  
this happens whenever the controller is activated by a change in driver attentiveness level, finds the driver fully attentive, and therefore switches off any activated alerts, and selects the nominal driving speed. In the second, the controller is activated by a timer whenever the driver has not become fully attentive after a period of time since a previous controller action. In this situation, the CTMC transitions from each state $(l,a,v,\overline{c})$ with $l\neq\mathsf{'A'}$ to the counterpart state $(l,a,v,c)$ in which the activated controller has the opportunity to switch on new alerts and/or to select a new speed level. 
    
\item[3)] \emph{Transitions corresponding to controller options.} When the controller is activated (by a change in driver attentiveness level or by the timer) and finds the driver to not be fully attentive, it has a choice of using any available combination of alerts and any speed level in order to make the driver attentive and to reduce risk. Thus, from each state $(l,a,v,c)\in S$ with $l\neq\mathsf{'A'}$, the CTMC can transition to any state $(l,a',v',c)\in S$. These controller options are indicated by dashed transitions in Figure~\ref{fig:CTMC}. In the general case from Section~\ref{subsect:problem}, there are $n$ driver attentiveness levels, $m$ alerts and $q$ speed levels, giving the controller $2^mq$ combinations of alerts and speed level options to choose from in each of the $(n-1)2^mq$ CTMC states in which $l\neq\mathsf{'A'}$. We encode the controller option for each of the $n-1$ driver attentiveness levels $l\in L\setminus\{\mathsf{'A'}\}$ and each of the $2^mq$ combinations of alerts and speed level $a\_v\in A\times V$ using a design-space parameter 
\begin{equation}
\label{eq:controller-options}
    \mathit{option}_{l,a\_v}\in\{0,1,\ldots,2^mq-1\}. 
\end{equation}
We have $(n-1)2^mq$ such parameters, and each assignment of values to these parameters defines a candidate deterministic controller\footnote{A deterministic controller is a controller which, for any state $s\in S$, performs the same action each time when state $s$ is reached.} solution for the driver attentiveness management problem. There are $(2^mq)^{(n-1)2^mq}$ candidate solutions in total, and thus $(2^22)^{(3-1)2^22}=8^{16}\approx 10^{14}$ candidate solutions for the problem instance encoded by the parametric CTMC from Figure~\ref{fig:CTMC}.
\squishend 

To complete the definition of the controller design space, we also need to specify its CTMC transition rates. The last two types of transitions described above correspond to controller actions. Therefore, their rates must reflect the mean operation time that the controller requires: (i)~to plan the new alerts and speed level when it is activated, and (ii)~to be activated by its timer. For instance, a controller operation time of 500ms gives a rate of 2s$^{-1}$. These rates are easy to determine, e.g., by worst-case execution time analysis of the controller code. In contrast, the rates for the first type of transition are much more difficult to determine because they encode the mean time taken by the driver to transition between attentiveness levels, for each combination of active alerts and speed levels. These rates must be estimated, e.g., by using data sources such as:
\squishlist
\item[1)] the numerous available studies and surveys of driver attentiveness, e.g.~\cite{korber2018have,lotz2018predicting,maia2013short,vanlaar2008fatigued};
\item[2)] additional data from controlled experiments with drivers of ALKS vehicles;
\item[3)] driver data collected during the actual driving of ALKS vehicles, e.g., by using a black-box solution similar to that already employed by many insurers of new drivers~\cite{kassem2008vehicle,kumar2018intelligent}, either across a fleet of vehicles or for a specific driver. 
\squishend
We note that using the last data source enables both (i)~the definition of personalised controller design spaces for each driver, and (ii)~the continual updating of these design spaces to support the runtime synthesis of new \acronym\ controllers when the transition rates for a driver change significantly~\cite{zhao2020interval}. 

\subsection{Synthesis of Pareto-optimal \acronym\ controllers \label{subsect:synthesis-Pareto}}

The synthesis of \acronym\ controllers solves the driver attentiveness management problem. For this purpose, we define the reward structures $r^\mathit{nuisance}$, $r^\mathit{progress}$ and $r^\mathit{risk}$ over the CTMCs from our controller design space. The definitions of these reward structures are directly based on the definitions of the three measures with the same names from Section~\ref{subsect:problem}. For instance, the state and transition reward functions for the first reward structure are given by $r^\mathit{nuisance}.r_1(l,a,v,c?)=\mathit{nuisance}(a)$ for any CTMC state $(l,a,v,c?)\in S$, and $r^\mathit{nuisance}.r_2(s_1,s_2)=0$ for any CTMC transition $(s_1,s_2)\in S\times S$, respectively. Given these reward structures, the driver attentiveness management problem for a journey of duration $T>0$ can be formalised as:

\vspace*{1.5mm}
\setlength{\leftskip}{0.5cm}
\noindent
\emph{Find the set of controller options~\eqref{eq:controller-options} whose associated CTMCs from the controller design space achieve Pareto-optimal trade-offs between minimising $R_{=?}^\mathit{nuisance}[C^{\leq T}]$, maximising $R^\mathit{progress}_{=?}[C^{\leq T}]$ and minimising $R_{=?}^\mathit{risk}[C^{\leq T}]$,}

\vspace*{1.5mm}
\setlength{\leftskip}{0cm}
\noindent
where the three CSL cumulative reward properties represent the overall nuisance, overall progress with the journey, and overall risk accrued over a journey of duration $T$, respectively.

We solve this problem using the search-based software engineering tool EvoChecker~\cite{gerasimou2015search,gerasimou2018synthesis}, which:
\squishlist
\item[1)] obtains the precise values of the three reward properties for any given CTMC from the controller design space using a probabilistic model checker (the tool can be configured to use PRISM~\cite{prism} or Storm~\cite{storm});
\item[2)] synthesises a close approximation of the Pareto-optimal set of controller options~\eqref{eq:controller-options} by using multi-objective genetic algorithm (MOGA) optimisation (the tool can be configured to work with any of the NGSA-II~\cite{deb2002fast}, SPEA2~\cite{zitzler2001spea2} or MOCell~\cite{nebro2009mocell} MOGAs).
\squishend

\begin{figure}
\centering
\includegraphics[width=\hsize]{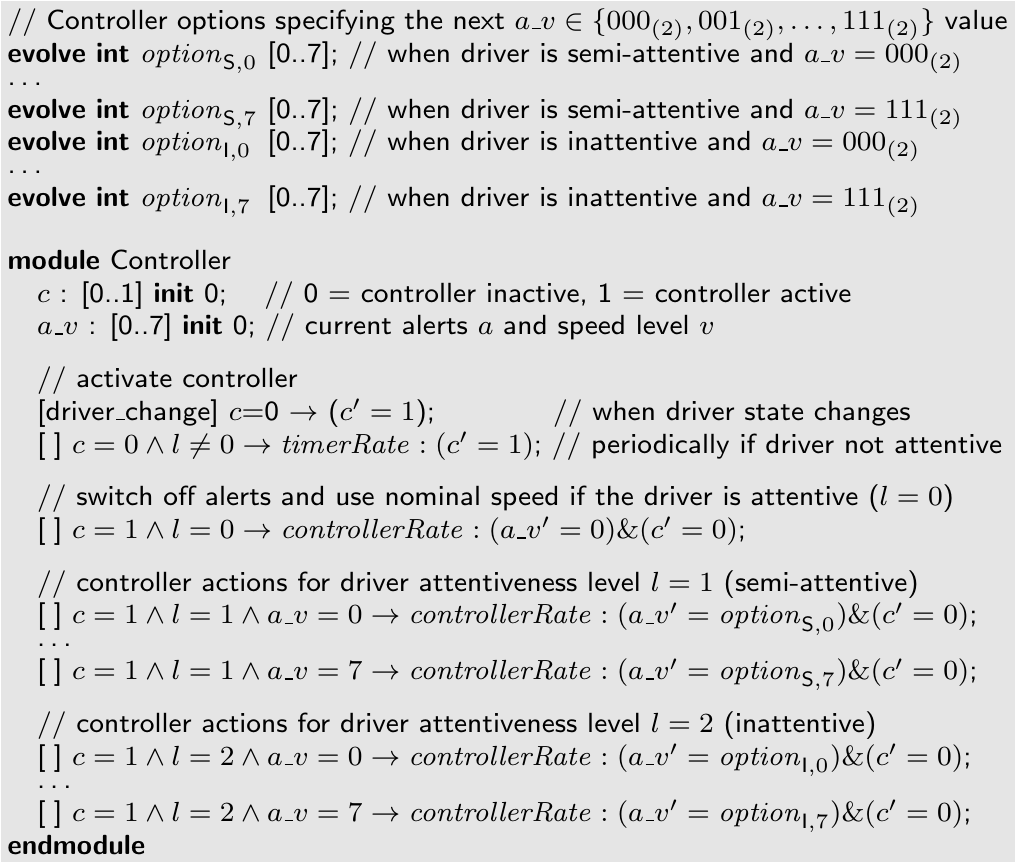}

\vspace*{-1mm}
\caption{Fragment of EvoChecker-encoded controller design space \label{fig:evochecker-model} for $m=2$ independent alerts and $q=2$ speed levels}

\vspace*{-3.5mm}
\end{figure}

\begin{figure}
\centering
\includegraphics[trim=43mm 8mm 22mm 30mm, clip, width=0.78\hsize]{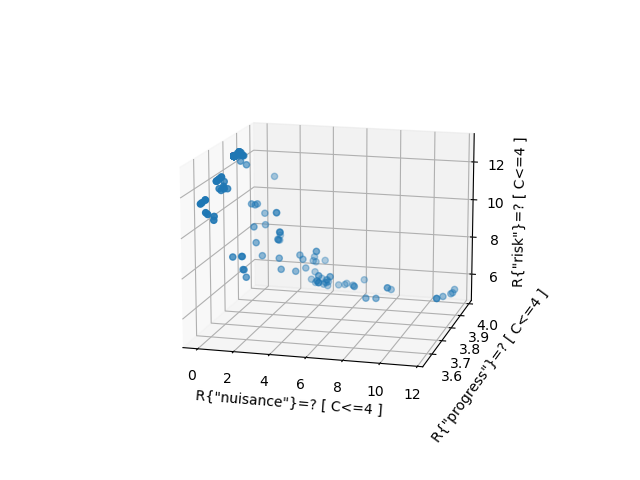}

\vspace*{-2mm}
\caption{Pareto front associated with the set of Pareto-optimal \acronym\ controllers for the controller design space instance from Figure~\ref{fig:CTMC}, as synthesised in 98.58s by EvoChecker configured to use Storm~\cite{storm} and NSGA-II~\cite{deb2002fast} (population size 7000 $\times$ 1000~iterations) and running on a 3.6GHz Intel Core i3 Mac OSX 10.14.6 Mac mini computer with 16 GB of memory \label{fig:pareto}}

\vspace*{-4mm}
\end{figure}

To this end, we supply EvoChecker with: (i)~our controller design space from Section~\ref{subsect:synthesis-design-space}, encoded in the high-level PRISM modelling language~\cite{prism} extended with EvoChecker constructs that we use to specify the possible values for the parameters~\eqref{eq:controller-options}; and (ii)~the three CSL reward properties specifying the optimisation objectives from our problem. Figure~\ref{fig:evochecker-model} shows how the controller design space from Figure~\ref{fig:CTMC} is expressed in this encoding. Due to space constraints, only a fragment of the encoding is shown, but we made the entire encoding (and the other artifacts from this section) available for inspection at \url{https://cutt.ly/SafeSCAD-SEAMS21}. Given the controller design space and the optimisation objectives, EvoChecker synthesises a close approximation of the Pareto-optimal set of \acronym\ controllers, and the Pareto front associated with this set. Figure~\ref{fig:pareto} shows the Pareto front obtained for the instance of the driver attentiveness management with the design space from Figure~\ref{fig:CTMC} and a driving time of $T=4$~hours. Each element of this Pareto front corresponds to a controller variant whose nuisance, risk and progress values from Figure~\ref{fig:pareto} were obtained by EvoChecker through formal verification using the Storm model checker.

\section{Related work}
\label{sec:relwork}

Human-machine interaction in driving automation includes the identification and
handling of control transitions between the driver and the vehicle
\cite{Walch2017-CarDriverHandovers, ISO21959} and, in support of that,
managing the driver's attentiveness and involvement necessary for such
transitions, taking into account the current traffic situation with
its potential adversity.

Recent research deals with managing such control transitions
\cite{ISO21959, Latotzki2020-HandoverProcedureDriver} or continuous
shared control \cite{Walch2017-CarDriverHandovers}, and includes 
inventions about control transitions that %
require but do
not manage driver attentiveness
\cite{Latotzki2020-HandoverProcedureDriver,
Martinez2013-Collisionmitigationsystems,
  Hoye2016-Determiningdriverengagement}.
Similarly, an earlier invention
\cite{Kopf2004-Systemsmethodsevaluating} focuses on classifying the
driver state and on technologies for situation-adapted collision
avoidance by combining braking and driver warnings.
A range of experiments \cite{Gold2013-Partiallyautomateddriving,
  Biondi2017-Advanceddriverassistance,
  Trimble2015-Humanfactorsevaluation} investigate parameters of the
driver state and behaviour (e.g.~response times, drowsiness, influence
of traffic density and driver workload), however, with little data
about the time it takes drivers to become inattentive.

Overall, none of the works we found discusses how attentiveness
monitoring and control software can be automatically designed and
adapted for optimal safety and performance, and how such software can be
confidently assured to fulfil regulatory requirements \cite{ISO21448}.
In contrast, our \acronym\ approach focuses on the design and analysis of a MAPE control loop supported by such
software, assuming the availability of specific sensor technology for
estimating the driver state \cite{Gold2013-Partiallyautomateddriving}.
Moreover, we provide a generic method for defining the design space for this control software, and for its automated synthesis with probabilistic guarantees. Finally, our exhaustive formal verification approach based on probabilistic model checking is also an improvement over the purely testing-based approach that the \ac{SOTIF} standard recommends for the verification of
automated driving systems \cite[\S 10.3,
Table~5-7]{ISO21448}.

\section{Conclusion \label{sect:conclusion}}

We introduced a MAPE control loop for improving driver attentiveness in ADS-enhanced cars, and we described a novel method for the rigorous synthesis of the controller used in the planning stage of this MAPE loop. In the next stage of our project, we will complete the development of the DNN-based analysis module from Figure~\ref{fig:analysis} by integrating our project's DeepTake predictor of driver takeover behaviour~\cite{CHI-2021} and its verification results~\cite{Grese-et-al-2020} into the end-to-end solution presented in this paper. The complete solution will then be evaluated experimentally, in a study carried out using our driving simulator from~\cite{CHI-2021}. 

Additionally, we will assess whether the good EvoChecker scalability reported in~\cite{gerasimou2018synthesis} extends to our controller synthesis problem with larger numbers of alerts $m$ and speed levels $q$ (cf.~Section~\ref{subsect:problem}). Finally, we plan to explore: (i)~the use of personalised and adaptive controllers (as described in Section~\ref{subsect:synthesis-design-space}); (ii)~the use of the CTMC-refinement technique from~\cite{paterson2017accurate,paterson2018observation} to improve the accuracy of the \acronym\ controller design space; (iii)~the use of the robust synthesis techniques from~\cite{CALINESCU2018140,7930209} to generate controllers tolerant to variations in driver behaviour; and (iv)~the potential advantages of using probabilistic \acronym\ controllers, i.e., controllers for which the options~\eqref{eq:controller-options} are discrete probability distributions over the set of actions available to the controller.

\section*{Acknowledgements}
This project has received funding from the Assuring Autonomy International Programme project `Safety of shared control in autonomous driving' and the UKRI project EP/V026747/1 `Trustworthy Autonomous Systems Node in Resilience'.

\bibliographystyle{IEEEtran}
\bibliography{}

\end{document}